\documentclass{article}

\usepackage[final, nonatbib]{nips_2018}

\usepackage[utf8]{inputenc} 
\usepackage[T1]{fontenc}    
\usepackage{hyperref}       
\usepackage{url}            
\usepackage{booktabs}       
\usepackage{amsfonts}       
\usepackage{nicefrac}       
\usepackage{microtype}      
\usepackage{enumitem,mathrsfs,amsmath,amssymb, color, underscore, bm, listings, bbm, graphicx, caption, subcaption, float}
\usepackage{threeparttable}
\usepackage{appendix}

\usepackage{biblatex}
\title{Contextual Aware Joint Probability Model Towards Question Answering System}

\author{
  Liu Yang\\
  Department of Computer Science\\
  Stanford University\\
  \texttt{liuyeung@stanford.edu}\\
  \And
  Lijing Song\\
  Department of Computer Science\\
  Stanford University\\
  \texttt{lisasong@stanford.edu}\\
}

\begin{document}

\maketitle

\begin{abstract}
In this paper, we address the question answering challenge with the SQuAD 2.0 dataset. We design a model architecture which leverages BERT's \cite{devlin2018bert} capability of context-aware word embeddings and BiDAF's \cite{seo2016bidirectional} context interactive exploration mechanism. By integrating these two state-of-the-art architectures, our system tries to extract the contextual word representation at word and character levels, for better comprehension of both question and context and their correlations. We also propose our original joint posterior probability predictor module and its associated loss functions. Our best model so far obtains F1 score of \boldsymbol{$75.842\%$} and EM score of \boldsymbol{$72.24\%$} on the test PCE leaderboad.
\end{abstract}

\section{Introduction}
With increasing popularity of intelligent mobile devices like smartphones, Google Assistant, and Alexa, machine question answering is one of the most popular field of deep learning research. SQuAD 2.0 \cite{rajpurkar2018know} challenges a question answering system in two respects: first, to predict whether a question is answerable based on a given context; second, to find a span of answer in the context if the question is predicted as answerable. It augments the 100,000 questions in SQuAD 1.1 \cite{rajpurkar2016squad} with 50,000 more unanswerable questions written in purpose to look like answerable ones. Because of the existence of those unanswerable questions which account for \nicefrac{1}{3} of the train set and \nicefrac{1}{2} of the dev and test sets, we design an elaborate joint probability predictor layer on top of our BERT-BiDAF architecture to solve the answerable classification and span prediction problem in a probability inference way. 

There are many efforts tackling the question answer systems and some work already achieve human level performance. However, common practice of these works actually base on several weird assumptions which we argue that will introduce tough constraints to the model and limit the model's representation capability. 

In this paper, we explicitly point out these weird assumptions and give a thorough discussion about their weakness. We propose our solution to model the question answer problem in a brand new joint probability way. Given a specific (question, context) pair, we try to make the model to learn the joint posterior distribution of the binary answerable or not variable and the start/end span variables. We propose a BERT-BiDAF \cite{devlin2018bert}\cite{seo2016bidirectional} hybrid model  to capture the question aware context information at both character and word levels with the expectation to gather strong signals to help our joint probability model make decisions. 

\section{Related Work}
On SQuAD leaderboards, all the top works apply BERT word embedding layer. As a pre-trained bidirectional transformer language model released in late 2018, BERT \cite{devlin2018bert} is highly respectable to produce context-aware word representation. Adding n-gram masking and synthetic self-training techniques onto the BERT framework, the current state-of-art model has achieved near human F1 and EM accuracy with ensembling. On top of BERT emdedding layer, we apply bidirectional attention mechanism, because it co-attends to both paragraphs and questions simultaneously and is used by all the top-ranking models for SQuAD. Among them, BiDAF \cite{seo2016bidirectional} is one of the first and most important models. The central idea behind BiDAF and its variants \cite{peters2018deep} is the Attention Flow layer, that generates a question-aware context representation by incorporating both context-to-question and question-to-context attentions. 

Other competitive approaches to question answering include QANet \cite{yu2018qanet} and Attention-over-Attention (AoA) \cite{cui2016attention}. QANet speeds up training by excluding RNNs and using only convolution and self-attention, where convolution models local interactions and self-attention models global interactions. AoA models the importance distribution of primary attentions by stacking additional attentions on top of the primary attentions.

\section{Approach} \label{sec-approach}
In this section, we formally define the terminology "context" as the paragraph in which we want to find the answer while "question" as it literally is. We propose the Joint BERT-BiDAF Question Answering neural model, Fig.1 illustrates its overall architecture. We keep the "Attention Flow layer" of BiDAF \cite{seo2016bidirectional} to produce the question aware context representation at character level. As BERT \cite{devlin2018bert} can directly provide question aware context at word level, we simply combine the two features together and feed them into BiDAF's "Modeling layer" \cite{seo2016bidirectional} with the expectation to obtain mutually enhanced context representation (mutual awareness of character/word features, mutual positional interactions, etc.). Intuitively, the word piece strategy of BERT \cite{devlin2018bert} doesn't split the normal words, thus including character embedding can be a good supplement. 

In Fig.\ref{fig-architecture}, the top layer is our proposed joint probability prediction structure. Unlike the common practice of inserting a "no-answer" token (e.g baseline and \href{https://github.com/huggingface/pytorch-pretrained-BERT/blob/master/examples/run_squad.py}{Bert squad}) and making the start/end position of the no-answer case converge to that special token, we try to model the joint distribution of the binary variable \(A\) (has/no answer), the N (context length) value variables \(X_{1}\) (start) and \(X_{2}\) (end) in a more natural way. We shall give a thorough discussion about this structure in section.\ref{jointprobsec}.

The purpose of introducing the new variable \(A\) is to make the model align to the causality that, given a specific question and context, the existence of a valid <start, end> span is depend on the answerable property other than the other way around. It is a big change which result in that we can't unify the no-answer loss under the general position entropy loss \cite{devlin2018bert}\cite{seo2016bidirectional}, we must design new loss functions which align to our joint probability structure. Loss function is essential since it exactly determine every step of the learning update. It is quite trick but interesting to try to figure out a loss function which value indeed reflect the model's performance while consistent with the model's design philosophy. We shall discuss our loss function exploration in details in section.\ref{lossfuncsec}. 

What should be emphasized is that we don't want to simply reproduce BERT's success in question answer problem on SQuAD 2.0 \cite{devlin2018bert}. The main difference between our solution and BERT's QA solution \footnote{\url{ https://github.com/huggingface/pytorch-pretrained-BERT/blob/master/examples/run_squad.py}} is BERT use representations of both question and context to vote for the span prediction while we only use that of the context. What we really want to do is to verify our original ideas, to test how well BERT can let the context "aware" the existence of question and to see whether the two embeddings at different granularity can achieve mutual enhancement. 

This section is scheduled in the following way: in section\ref{arch}, we discuss about our \textit{original} model architecture in general and our effort to implement this architecture. We go through the essential layers and their back-end techniques but leave the BiDAF\cite{seo2016bidirectional} details for reference reading. Especially in section\ref{jointprobsec}, we shall focus on our \textit{original} joint probability predictor and how we tackle the answerable problem and the span prediction problem. In section\ref{lossfuncsec}, we shall show several our \textit{original} loss functions and their considerations. In section\ref{baseline}, we introduce the baseline we use and in section\ref{workload}, we briefly summarize the workload of our project. Please keep in mind that we define our terminology incrementally, once a symbol has been defined, it will be used throughout the rest of paper.

\subsection{Model Architecture} \label{arch}
\begin{figure}[H]
\centering
\includegraphics[width=\linewidth]{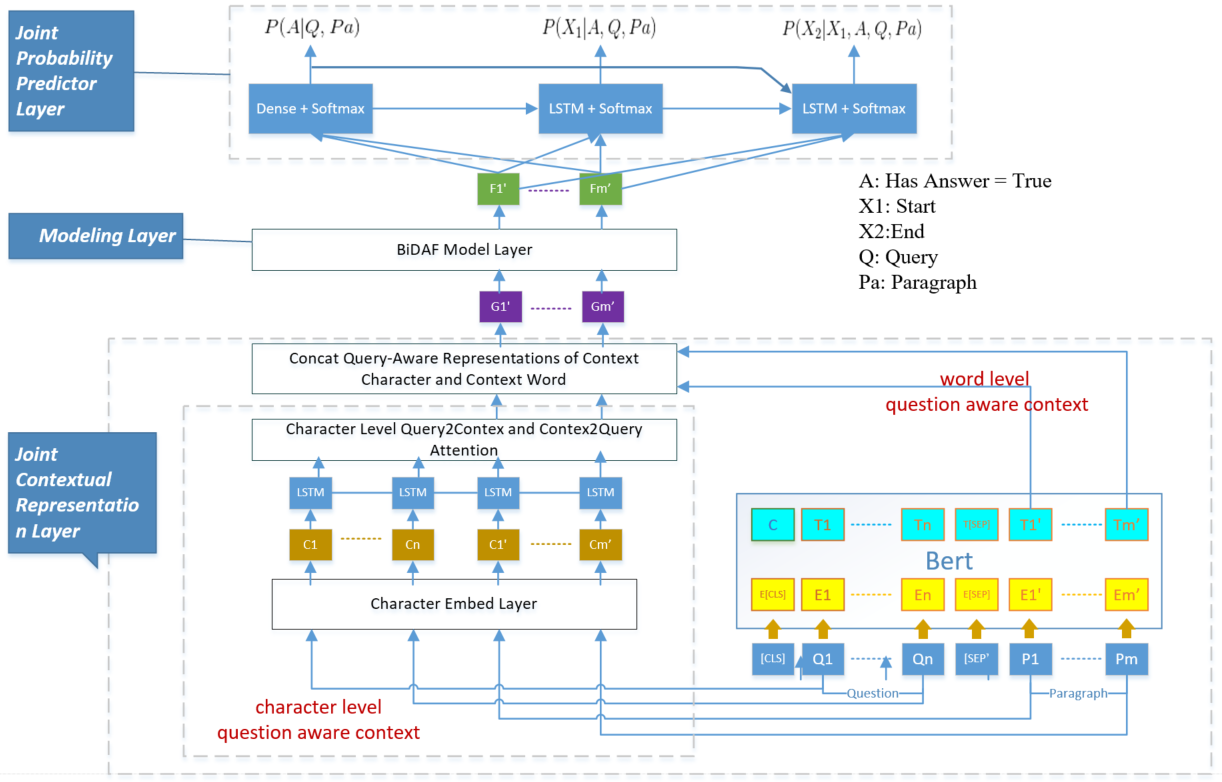}
\caption{The Joint BERT-BiDAF Question Answering Model}
\label{fig-architecture}
\end{figure}

As illustrated in Fig. \ref{fig-architecture}, we keep the main framework of BiDAF \cite{seo2016bidirectional} but replace its GloVe \cite{pennington2014glove} word embedding with BERT contextual word embedding \cite{devlin2018bert} and stack joint probability predictor on top of it. There are 3 hyper-parameters that determine the computational complexity of our system and we formally define them to be \(d_{lstm}, d_{char\_emb}\) and \(d_{bert}\). \(d_{lstm}\) is a uniform hidden dimension  for all LSTM components \cite{hochreiter1997lstm}. \(d_{char\_emb}\) and \(d_{bert}\) are the dimension of the character word embedding (output of character level CNN \cite{zhang2015character}) and the hidden dimension of the \(BERT_{BASE}\)\cite{devlin2018bert}, respectively.

\subsubsection{Embedding Layer}
Our embedding layer is built based on \href{https://github.com/huggingface/pytorch-pretrained-BERT/blob/master/README.md}{BERT base uncased tokenizer}.
\begin{enumerate}[label=(\roman*)]
    \item \textbf{Word Level Embedding}: More precisely, this is token level embedding. Yet BERT doesn't split common words thus it is still can be treated as word representation.
    \item \textbf{Character Level Embedding}: BERT split special words such as name entity into pieces started with "\#\#", e.g "Giselle" to ['gi', '\#\#selle']. However, each piece is still a character string from which a \(d_{char\_emb}\) dimensional embedding can be generated with character CNN \cite{zhang2015character}. Note that "\#\#" in word pieces is ignored.  
\end{enumerate}

\subsubsection{Joint Contextual Representation Layer }
In this layer, we denote the lengths of question and context as \(L_{q}\) and \(L_{c}\) respectively, the character level representations of question and context are \(R_{cq} \in R^{L_{q}\times d_{char\_emb}}\) and \(R_{cc} \in R^{L_{c}\times d_{char\_emb}}\) respectively. We apply the \textit{"Attention Flow Layer"} of BiDAF\cite{seo2016bidirectional} to obtain \(T_{Char} \in R^{8d_{char\_emb} \times L_{c}}\) as context representation conditioned on question at character level. 

We simply keep the context part from the final encoder layer of \(BERT_{BASE}\) to obtain \(T_{Word} \in R^{d_{bert} \times L_{c}}\) as context representation conditioned on question at word level. We concatenate the two representations together to produce the joint contextual representation \(G_{ctx} = [T_{Word}, T_{Char}] \in R^{(8d_{char\_emb} + d_{bert}) \times L_{c}}\). 

\subsubsection{Modeling Layer}
We keep the \textit{"BiDAF Modeling Layer"}\cite{seo2016bidirectional} for two reasons.
\begin{enumerate}[label=(\roman*).]
    \item Although BERT word \(T_{Word}\) can be treat it as positional aware representation but the character feature \(T_{Char}\) can not. We expect \(T_{Word}\) propagates the positional information to \(T_{Char}\) in the mixture process in this layer.
    \item We want the fine granularity information captured at character level propagates to \(T_{Word}\). Actually we expect the two level representations aware the mutual existence and learn how to cooperate to produce the augmented feature. 
\end{enumerate}
The output of the Modeling Layer is \(F \in R^{(2d_{lstm} + 8d_{char\_emb} + d_{bert}) \times L_{c} }\)

\subsubsection{Joint Probability Predictor Layer} \label{jointprobsec}
There are 3 components in our joint probability predictor layer which are responsible to compute the \(P(A)\), \(P(X_{1} | A)\) (start position) and \(P(X_{2} | A, X_{1})\) (end position) respectively. We propose this structure based on the observation that common practices highly rely on the special "sentry" token and actually make a solid assumption that \(P(X_{1} = 0, X_{1} = 0 | A = 0)\). We argue that assumption like this introduces a tough constraint and the effort the model spends on push the \((X_{1}, x_{2})\) to \((0,0)\) will definitely impact the model's internal mechanism to generate \(X_{1}, X_{2}\) condition on \(A = 1\).

Our method gives up the special "sentry" token, we choose the natural domain of \(X_{1}, X_{2}\) to be \(\{0,1,\cdots, L_{c}-1\}^{2}\), and relax the \(P(X_{1} = 0, X_{2} = 0| A = 0)\) constraint to \(P(X_{1} = i, X_{2} = j | A = 0), \forall i > j\), that is, the predicted start position should be \textit{larger} than the predicted end position condition on \textit{no-answer}. The fact \(P(X_{1}=i, X_{2} = j | A = 1), \forall i \leq j\) becomes a natural extension of our concept. Thus we claim that our method does not introduce any inherent contradiction and any bias assumption.  

The connection of the 3 predictors are consistent with the chain rule, the \(P(X_{1} | A)\) predictor relies on the \(P(A)\) predictor, while the \(P(X_{2} | A, X_{1})\) predictor relies on the other two both. Thus the structure is expected to simulate the joint probability \(P(A, X_{1}, X_{2}) = P(A)P(X_{1} | A)P(X_{2} | A, X_{1})\). 

All the 3 predictors take the output \(F \in R^{(2d_{lstm} + 8d_{char\_emb} + d_{bert}) \times L_{c} }\) of the \textit{"Modeling Layer"} as input. For simplicity, we define \(f = 2d_{lstm} + 8d_{char\_emb} + d_{bert}\).\newline

\textbf{I. \(P(A)\) Predictor}\\
In order to compute the binary classification logits, we use a \textit{self attention} mechanism by setting a learn-able parameter \(w_{A} \in R^{f}\), the attention vector \(att_{A} \in R^{L_{c}}\) is computed by Eq.[\ref{eq:probA}] 
\begin{equation} \label{eq:probA}
    att_{A}[i] = \frac{exp(w_{A}^{T} F[:,i])}{\sum_{k=0}^{L_{c}-1}exp(w_{A}^{T} F[:,k])}, i \in \{0,1,\cdots, L_{c}-1]\}
\end{equation}
We compute the context summary vector by \(S_{A} = F att_{A} \in R^{f}\). Then we use a learn-able matrix \(W_{logit} \in R^{2 \times f}\) to compute the classification logits \(l_{A} = W_{logits}S_{A} \in R^{2}\), thus we obtain \(P(A)\) as Eq.[\ref{eq:A01}]
\begin{equation} \label{eq:A01}
    P(A = 0) = \frac{exp(l_{A}[0])}{exp(l_{A}[0]) + exp(l_{A}[1])}, 
    P(A=1) = \frac{exp(l_{A}[1])}{exp(l_{A}[0]) + exp(l_{A}[1])}
\end{equation}

In order to propagate the decision of \(P(A)\) predictor to \(P(X_{1} | A)\) predictor, we use a learn-able matrix \(W_{prop}^{A} \in R^{2d_{lstm} \times f}\) to generate a tuple \((h_{A}, c_{A}) = W_{prop}^{A}S_{A}\) to initialize the LSTM layer of the \(P(X_{1} | A)\) predictor.\newline

\textbf{II. \(P(X_{1} | A)\) Predictor}\\
By the bidirectional LSTM layer in \(P(X_{1} | A)\) predictor, we obtain \(M_{1} = LSTM(F, (h_{A}, c_{A})) \in R^{2d_{lstm}\times L_{c}}\). We use a learn-able vector \(w_{1} \in R^{2d_{lstm}}\) to compute the distribution of \(X_{1}\) in Eq.[\ref{eq:probx1}]
\begin{equation} \label{eq:probx1}
    P(X_{1} = i | A) = \frac{exp(w_{1}^{T} M_{1}[:,i])}{\sum_{k=0}^{L_{c}-1}exp(w_{1}^{T} M_{1}[:,k])}, i \in \{0,1, \cdots, L_{c}-1\})
\end{equation}
We use the \(P(X_{1}| A) \in R^{L_{c}}\) itself as attention to generate the context summary \(S_{1} = MP(X_{1} | A) \in R^{2d_{lstm}}\). We concatenate the two context summaries obtained so far as \(S_{joint} = [S_{A}, S_{1}]\) and propagate the decisions of predictors \(P(A)\) and \(P(X_{1}|A)\) to predictor \(P(X_{2} | A, X_{1})\) by initialize its bidirectional LSTM by \((h_{1}, c_{1}) = W_{prop}^{1}S_{joint}\). Here \(W_{prop}^{1} \in R^{2d_{lstm} \times (f + 2d_{lstm})}\) is another learn-able matrix parameter in this predictor.\newline

\textbf{III. \(P(X_{2} | A, X_{1})\) Predictor} \\
By the bidirectional LSTM layer in \(P(X_{2} | A, X_{1})\) predictor, we obtain \(M_{2} = LSTM(F, (h_{1}, c_{1})) \in R^{2d_{lstm}\times L_{c}}\). We use a learn-able vector \(w_{2} \in R^{2d_{lstm}}\) to compute the distribution of \(X_{2}\) in Eq.[\ref{eq:probx2}]
\begin{equation} \label{eq:probx2}
    P(X_{2} = i | A, X_{1}) = \frac{exp(w_{2}^{T} M_{2}[:,i])}{\sum_{k=0}^{L_{c}-1}exp(w_{2}^{T} M_{2}[:,k])}, i \in \{0,1,\cdots, L_{c}-1\})
\end{equation}\newline

\textbf{IV. How to Predict}\\
For a specific instance \(\mathcal{D}[m]\), from Eq. \ref{eq:A01}, Eq. \ref{eq:probx1} and Eq. \ref{eq:probx2} we have the joint posterior probability as Eq. \ref{eq:joint}:
\begin{equation}\label{eq:joint}
  \begin{split}
    P(A = a, X_{1} = i, X_{2} =j| \mathcal{D}[m]) =
    P(A=a|\mathcal{D}[m])P(X_{1}=i | A=a, \mathcal{D}[m])P(X_{2}=j|A=a,X_{1}=i, \mathcal{D}[m]) \\
    a \in \{0, 1\}, (i,j) \in \{0,1,\cdots L_{c}-1\}^{2}
  \end{split}
\end{equation}
We find the maximal negative likelihood \(p_{0} = max_{i>j}\{P(A=0, X_{1}=i, X_{2} = j) | \mathcal{D}\}\) and the maximal positive likelihood \(p_{1} = max_{i\leq j}\{P(A=1, X_{1}=i, X_{2}=j | \mathcal{D})\}\) . If \(p_{1} > p_{0}\) then we predict the instance \(\mathcal{D}[m]\) has answer, otherwise, \(\mathcal{D}[m]\) has \textbf{no} answer. If \(\mathcal{D}\) predicted as has answer, then \((start, end) = argmax_{i\leq j}\{P(A=1, X_{1}=i, X_{2}=j | \mathcal{D}[m])\}\). 

\subsection{Loss Function} \label{lossfuncsec}
Assume for a specific instance \(\mathcal{D}[m]\), the ground truth is $(A, X_{1}, X_{2}) = ( a, i, j)$. We propose our first loss function of our joint probability predictor in Eq. \ref{loss1}. Intuitively, in addition to the normal binary cross entropy, when an instance \(\mathcal{D}[m]\) has no answer, we want the probability concentrates on the maximum likelihood estimation (MLE); when an instance has answer, we just punish the position loss as common practice.  
\begin{equation}\label{loss1}
    \begin{split} 
        loss_{1}(\mathcal{D}[m]) =& -log(P(A=a | \mathcal{D}[m])) + \mathbbm{1}(a=0)\{-log(max_{i>j}\{P(A=0, X_{1}=i, X_{2}=j|\mathcal{D}[m])\})\}\\
        & \quad + \mathbbm{1}(a=1)\{-log(P(X_{1}=i|A=1, \mathcal{D}[m]) - log(X_{2}=j | A=1, X_{1}=i, \mathcal{D}[m])\}
    \end{split}
\end{equation}
For our second loss function, we introduce the distribution \(\mathcal{U}(i,j)\) as Eq. \ref{uniform}. \(\mathcal{U}(i,j)\) is a partial uniform distribution whose probability mass concentrates in the confidence area of \(P(A=0|\mathcal{D}[m])\).
\begin{equation} \label{uniform}
    \mathcal{U}(i,j) = \begin{cases}
             \frac{2}{(L_{c}-1)L_{c}}, i > j\\
             0, i \leq j
             \end{cases}
\end{equation}
When an instance has no answer, we want the model produces a distribution \(P(X_{1}, X_{2} | \mathcal{D}[m])\) close to \(\mathcal{U}\). With the Kullback-Leibler divergence, we give our \(loss_{2}\) as Eq. \ref{kl}:
\begin{equation} \label{kl}
    \begin{split} 
        loss_{2}(\mathcal{D}[m]) =& -log(P(A=a | \mathcal{D}[m])) + \mathbbm{1}(a=0)\{KL(\mathcal{U}(X_{1},X_{2})||P(X_{1},X_{2}|\mathcal{D}[m]))\}\\
        & \quad + \mathbbm{1}(a=1)\{-log(P(X_{1}=i|A=1, \mathcal{D}[m]) - log(X_{2}=j | A=1, X_{1}=i, \mathcal{D}[m])\}
    \end{split}
\end{equation}

\subsection{Baseline} \label{baseline}
Our baseline is a modified BiDAF \cite{seo2016bidirectional} model with word-level embedding provided by cs224n teaching staff \footnote{Chapter 4 in DFP Handout and repository cloned from \url{https://github.com/chrischute/squad.git}}. With the default hyper-parameters, e.g learning rate, decay weight etc. after $30$ epochs of training we get the performance metrics of baseline listed in Table \ref{table:performance}.

\subsection{Workload} \label{workload}
In our project, we install pre-trained BERT from huggingface\footnote{\url{https://github.com/huggingface/pytorch-pretrained-BERT}}. When reusing the training framework of the baseline, however, we totally change its tokenization logic in order to integrate BERT. We write a lot of such foundation code to make the two different system consistent. We write all other code to implement our proposed BERT-BiDAF hybrid architecture, our joint probability predictor and loss functions. For ablation purpose, we also implement another answer/no-answer classifier as a candidate for comparison.  

\section{Experiments} \label{sec-experiments}
\subsection{Dataset}
The dataset we use is Stanford Question Answering Dataset 2.0 (SQuAD 2.0) \cite{rajpurkar2018know}, a reading comprehension dataset boosted with additional $50,000$ questions which can not be answered based on the given context. The publicly available data are split into $129,941$, $6,078$, and $5,915$ examples acting as the train/dev/test datasets respectively. 

Performance is evaluated by two metrics: F1 and EM score. F1 is the harmonic average of precision and recall. EM (Exact Match) measures whether the system output matches the ground truth answer exactly, and is stricter than F1. AvNA (Answer vs. No Answer) is an auxiliary metric that measures how well a model works on labeling a question as has-answer or no-answer.

\subsection{Experimental Details}
In our experiment, we use \(BERT_{BASE}\) \cite{devlin2018bert} which configuration (hidden_size=768, intermediate_size=3072, layers=12). For the 3 model complexity identifiers we defined in section\ref{arch}, we have \(d_{bert} = 768\). In order to determine \(d_{char\_emb}\) and \(d_{lstm}\) and the learning rate, we do grid search in the range \(lr \in \{ke^{-5} | k\in\{2, 5, 6, 7, 10\}\}\), \(d_{char\_dim} \in \{16, 32, 64, 80\}, d_{lstm} \in \{32,64,128\}\). We have 60 setting here and for each setting we run 1 epoch on a Nvidia Titan V GPU and sort the settings by their F1 scores. It takes us about 4 days to get a sorting list and we then fine tune the model with 3 epochs by trying the high scored settings in order. Finally, we obtain our best model with \(lr = 5e^{-5}, d_{char\_emb} = 16, d_{lstm} = 64\).

\begin{figure}[H]
\centering
\includegraphics[width=\linewidth]{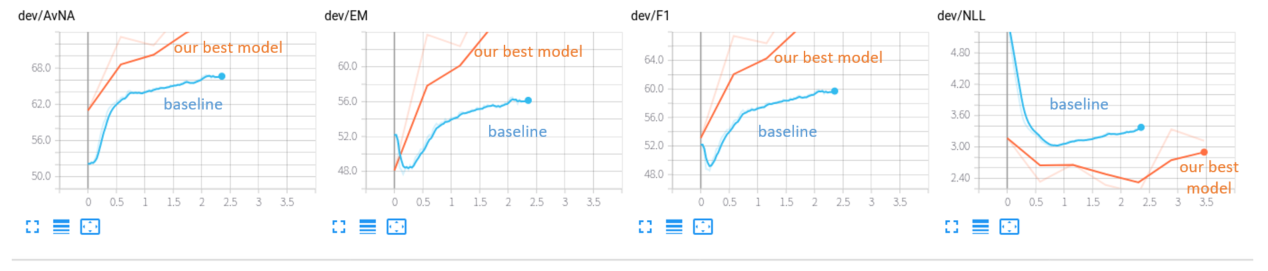}
\caption{Model Performance on dev-set During Training}
\label{fig-train}
\end{figure}

Fig.\ref{fig-train} depicts the performance of our model on dev dataset along with the training process going on [tensorboard visual output in 'Relative' mode]. The blue curves represent the baseline while the yellow ones represent our best model. For all the 3 metrics AvNA, EM and F1, our model beats the baseline by a significant margin. For the rightmost 'NNL' subplot, we can identify that the baseline tends to overfit the dataset in early stage of training, however, for our best model, the signals of overfitting appear much latter. 

We also conduct two experiments for ablation. In the first one, we set \(d_{char\_emb} = 2\), that actually makes the effect of character level embedding negligible. Then the experiment shows both F1 and EM scores punished by 2+ decrements. In the second one, because BERT\cite{devlin2018bert} claims that its hidden state of the final layer corresponding to '[CLS]' token can be used as the aggregated sequence representation for classification task, we simply use it for binary answerable classification. In this case, experiment shows AvNA decreases by 1.5 while EM and F1 both drop around 2, which indicates the advance of our output layer.

\subsection{Results}
We submitted our best model so far to the test PCE leaderboard and obtained the results \textbf{F1: 75.84, EM: 72.24}. However, in this section, in order to compare with other local implementation, we report the performance data on dev set in Table \ref{table:performance}. We list the performance data of the baseline, our best model, and the two ablation experiments. The performance differences have been described in the last section.

Although the overall performance differences are consistent to our arguments that character level embedding can help and that the joint probability predictor has decent performance, we believe the performance of our joint probability model can be improved by overcoming 2 difficulties. First, the overfitting problem. We note that the overfitting problem is significant in our system in the last epoch. Our best guess towards this problem is the unbalance parameter updating. As BERT uses a strategy which pushes learning rate to 0 when approaching to the end of training. Thus the gradient may vanish for BERT module but our upper stacked layers still keep learning, because they are closer to the gradient source (loss function). This unbalance learning makes the BERT module and the upper layer out of sync and the discrepancy within our system become bigger and bigger. Second, due to the limitation of the hardware, we can't run the size of batch as large as Google. Thus our parameter updating suffers from the variance problem which makes our system unstable.   

\begin{table} [H]
  \caption{Model Performance on Dev Set}
  \label{table:performance}
  \centering
  \begin{tabular}{llll}
    \toprule
    &\multicolumn{3}{c}{Metrics} \\
    \cmidrule(r){2-4}
    Model & F1 & EM & AvNA \\
    \midrule
    Baseline & 59.94 & 56.55 & 66.90 \\
    Our Best Model\ & 75.48 & 71.96 & 79.68 \\
    Without CharEmb & 73.25 & 69.87 & 78.01 \\
    \([CLS]\) AvNA Classifier & 73.49 & 69.76 & 78.23 \\
    Human \cite{rajpurkar2018know} & 89.45 & 86.83 & - \\
    \bottomrule
  \end{tabular}
\end{table}

\section{Qualitative Analysis of Model Errors} \label{sec-analysis}
\subsection{Error Type 1: Wrong Inference}
Wrong inference happens when the question and context pair is interpreted incorrectly by our model, and in some cases attention mechanism focuses on the wrong part of the context. This could happen because of underfitting in embedding or bidirectional attention, or when the sentence that includes the answer span is interpreted wrongly due to syntactic complication.

\textbf{I. False positive example of wrong inference} \newline
When reading the long list of Harvard's alumni, our system, same with original BERT, gets wrong on many of questions similar to the following ones. When the question asks for "actor starred in The Men in Black" (or "directer of Noah in 2014"), the system answers "Tatyana Ali" (or "Darren Aronofsky") simply because s/he is an "actor" (or "film director") according to the context, although "The Men in Black" or "Noah" is not even mentioned in the paragraph. Because of lack of well-trained embeddings of entities like film titles, the attention flow layer only focuses on key words such as "actor" and "director", and generates false positive results distracted by disturbance terms like "The Men in Black" or "Noah".

\textit{Question:} What actor starred in the Movie Saving Private Ryan? What actor starred in The Men in Black? \textit{(two questions)} \newline
\textit{Context:} ... comedian, television show host and writer Conan O'Brien; actors Tatyana Ali, Nestor Carbonell, Matt Damon, ... (\ref{a-para-alumni}. Harvard University in Appx. \ref{a-paragraph}) \newline
\textit{Ground Truth:} <No Answer>; <No Answer> \textit{(respectively)} \newline
\textit{Prediction:} Tatyana Ali; Tatyana Ali \textit{(respectively)}

\textbf{II. False negative example of wrong inference} \newline
In the context, there is no appearance of the key word "status", but the question is looking for such an answer. Our model fails to draw the correlation between the phrase and the word "status". This implies insufficiency either in the word embedding of "status" or in the bidirectional attention.

\textit{Question:} What status has the Brotherhood obtained in the Islamic world? \newline
\textit{Context:} Despite periodic repression, the Brotherhood has become one of the most influential movements in the Islamic world ... (\ref{a-para-status}. Islamism in Appx. \ref{a-paragraph}) \newline
\textit{Ground Truth:} one of the most influential movements \newline
\textit{Prediction:} N/A

\subsection{Error Type 2: Imprecise Answer Boundaries}
The model points to the right context part but results in imprecise answer span. More sophisticated techniques to predict start/end positions will help.

\textbf{I. Example of imprecise answer boundaries} \newline
    The prediction gives an unnecessary word "that".

\textit{Question:} What was Shrewsbury's conclusion? \newline
\textit{Context:} ... Shrewsbury in 1970, who noted that the reported rates of mortality in rural areas during the 14th-century pandemic were inconsistent with the modern bubonic plague, leading him to conclude that contemporary accounts were exaggerations... (\ref{a-para-conclude}. Black Death in Appx. \ref{a-paragraph}) \newline
\textit{Ground Truth:} contemporary accounts were exaggerations \newline
\textit{Prediction:} that contemporary accounts were exaggerations

\subsection{Error Type 3: Confusing Context}
Meanwhile, our system also makes mistakes that are prone to human readers. Sometimes we don't even agree with the ground truth answers. This error happens because of ambiguous context and tricky questions.

\textbf{I. Example of confusing context with imprecise answer boundaries} \newline
We and our model don't agree with the 3 ground truth answers and suspect it makes a coordination attachment error. The context is confusing with four and's in a single sentence.
    
\textit{Question:} What gained ground when Arab nationalism suffered? \newline
\textit{Context:} ... Arab nationalism suffered, and different democratic and anti-democratic Islamist movements inspired by Maududi and Sayyid Qutb gained ground. (\ref{a-para-inspire}. Islamism in Appx. \ref{a-paragraph}) \newline
\textit{Ground Truth:} anti-democratic Islamist movements, anti-democratic Islamist movements inspired by Maududi and Sayyid Qutb, anti-democratic Islamist movements \newline
\textit{Prediction:} different democratic and anti-democratic Islamist movements

\textbf{II. Example of confusing context without any error} \newline
Here is another example in which we don't agree with the ground truth answer, although our prediction is exactly the same as the ground truth. According to the context (and China's history), Kusala is Tugh Temür's brother, and was thought to be killed by El Temür; Tugh Temür stayed alive after his brother's death. Based on this comprehension, the right answer should be "No Answer" because Tugh Temür was never killed in the paragraph.

\textit{Question:} Who was thought to have killed Tugh Temur? \newline
\textit{Context:} ... Tugh Temür abdicated in favour of his brother Kusala, who was backed by Chagatai Khan Eljigidey, and announced Khanbaliq's intent to welcome him. However, Kusala suddenly died only four days after a banquet with Tugh Temür. He was supposedly killed with poison by El Temür, and Tugh Temür then remounted the throne. ... (\ref{a-para-kill}. Yuan Dynasty in Appx. \ref{a-paragraph}) \newline
\textit{Ground Truth:} El Temür \textit{(should be "No Answer")} \newline
\textit{Prediction:}  El Temür \newline

\section{Conclusion} \label{sec-conclusion}
In this paper, we propose our BERT-BiDAF hybrid architecture, thoroughly explain the intuition, and show the theoretical derivation of our joint probability predictor and its loss functions. Furthermore, we demonstrate that our ideas can obtain decent performance on a difficult and highly reputable problem (SQuAD 2.0) and verify our arguments by ablation experiments. We learn much about model design and how to conduct a effective learning. We identify that the overfitting and high variance problems are the two main difficulties which prohibit us form further progress. Our next step is to dig deeply into the difficulties we are facing and may take question representation into consideration. 

\subsubsection*{Acknowledgments}
Thanks to all the CS224 teaching staff for guidance and Microsoft Azure for their great generosity.

\printbibliography
\begin{appendices}

\section{Appendix: Full Paragraphs in Analysis} \label{a-paragraph}
\begin{enumerate}
    \item \textbf{Harvard University:} Other: Civil rights leader W. E. B. Du Bois; philosopher Henry David Thoreau; authors Ralph Waldo Emerson and William S. Burroughs; educators Werner Baer, Harlan Hanson; poets Wallace Stevens, T. S. Eliot and E. E. Cummings; conductor Leonard Bernstein; cellist Yo Yo Ma; pianist and composer Charlie Albright; composer John Alden Carpenter; comedian, television show host and writer Conan O'Brien; actors Tatyana Ali, Nestor Carbonell, Matt Damon, Fred Gwynne, Hill Harper, Rashida Jones, Tommy Lee Jones, Ashley Judd, Jack Lemmon, Natalie Portman, Mira Sorvino, Elisabeth Shue, and Scottie Thompson; film directors Darren Aronofsky, Terrence Malick, Mira Nair, and Whit Stillman; architect Philip Johnson; musicians Rivers Cuomo, Tom Morello, and Gram Parsons; musician, producer and composer Ryan Leslie; serial killer Ted Kaczynski; programmer and activist Richard Stallman; NFL quarterback Ryan Fitzpatrick; NFL center Matt Birk; NBA player Jeremy Lin; US Ski Team skier Ryan Max Riley; physician Sachin H. Jain; physicist J. Robert Oppenheimer; computer pioneer and inventor An Wang; Tibetologist George de Roerich; and Marshall Admiral Isoroku Yamamoto. \label{a-para-alumni}
    \item \textbf{Islamism:} Despite periodic repression, the Brotherhood has become one of the most influential movements in the Islamic world, particularly in the Arab world. For many years it was described as "semi-legal" and was the only opposition group in Egypt able to field candidates during elections. In the Egyptian parliamentary election, 2011–2012, the political parties identified as "Islamist" (the Brotherhood's Freedom and Justice Party, Salafi Al-Nour Party and liberal Islamist Al-Wasat Party) won 75\% of the total seats. Mohamed Morsi, an Islamist democrat of Muslim Brotherhood, was the first democratically elected president of Egypt. He was deposed during the 2013 Egyptian coup d'état. \label{a-para-status}
    \item \textbf{Black Death:} The plague theory was first significantly challenged by the work of British bacteriologist J. F. D. Shrewsbury in 1970, who noted that the reported rates of mortality in rural areas during the 14th-century pandemic were inconsistent with the modern bubonic plague, leading him to conclude that contemporary accounts were exaggerations. In 1984 zoologist Graham Twigg produced the first major work to challenge the bubonic plague theory directly, and his doubts about the identity of the Black Death have been taken up by a number of authors, including Samuel K. Cohn, Jr. (2002), David Herlihy (1997), and Susan Scott and Christopher Duncan (2001). \label{a-para-conclude}
    \item \textbf{Islamism:} The quick and decisive defeat of the Arab troops during the Six-Day War by Israeli troops constituted a pivotal event in the Arab Muslim world. The defeat along with economic stagnation in the defeated countries, was blamed on the secular Arab nationalism of the ruling regimes. A steep and steady decline in the popularity and credibility of secular, socialist and nationalist politics ensued. Ba'athism, Arab socialism, and Arab nationalism suffered, and different democratic and anti-democratic Islamist movements inspired by Maududi and Sayyid Qutb gained ground. \label{a-para-inspire}
    \item \textbf{Yuan Dynasty:} When Yesün Temür died in Shangdu in 1328, Tugh Temür was recalled to Khanbaliq by the Qipchaq commander El Temür. He was installed as the emperor (Emperor Wenzong) in Khanbaliq, while Yesün Temür's son Ragibagh succeeded to the throne in Shangdu with the support of Yesün Temür's favorite retainer Dawlat Shah. Gaining support from princes and officers in Northern China and some other parts of the dynasty, Khanbaliq-based Tugh Temür eventually won the civil war against Ragibagh known as the War of the Two Capitals. Afterwards, Tugh Temür abdicated in favour of his brother Kusala, who was backed by Chagatai Khan Eljigidey, and announced Khanbaliq's intent to welcome him. However, Kusala suddenly died only four days after a banquet with Tugh Temür. He was supposedly killed with poison by El Temür, and Tugh Temür then remounted the throne. Tugh Temür also managed to send delegates to the western Mongol khanates such as Golden Horde and Ilkhanate to be accepted as the suzerain of Mongol world. However, he was mainly a puppet of the powerful official El Temür during his latter three-year reign. El Temür purged pro-Kusala officials and brought power to warlords, whose despotic rule clearly marked the decline of the dynasty. \label{a-para-kill}
\end{enumerate}
\end{appendices}

\end{document}